\documentclass[10pt,twocolumn,letterpaper]{article}

\usepackage{cvpr}              

\usepackage{graphicx}
\usepackage{amsmath}
\usepackage{amssymb}
\usepackage{booktabs}

\usepackage[pagebackref,breaklinks,colorlinks]{hyperref}

\usepackage[capitalize]{cleveref}
\crefname{section}{Sec.}{Secs.}
\Crefname{section}{Section}{Sections}
\Crefname{table}{Table}{Tables}
\crefname{table}{Tab.}{Tabs.}

\def\cvprPaperID{*****} 
\def\confName{CVPR}
\def\confYear{2022}

\begin{document}

\title{Image Semantic Relation Generation}

\author{Mingzhe Du\\
School of Computer Science and Engineering, NTU\\
50 Nanyang Ave, Singapore 639798\\
{\tt\small mingzhe001@e.ntu.edu.sg}
}
\maketitle

\begin{abstract}
    Scene graphs provide structured semantic understanding beyond images. For downstream tasks, such as image retrieval, visual question answering, visual relationship detection, and even autonomous vehicle technology, scene graphs can not only distil complex image information but also correct the bias of visual models using semantic-level relations, which has broad application prospects. However, the heavy labour cost of constructing graph annotations may hinder the application of PSG in practical scenarios. Inspired by the observation that people usually identify the subject and object first and then determine the relationship between them, we proposed to decouple the scene graphs generation task into two sub-tasks: 1) an image segmentation task to pick up the qualified objects. 2) a restricted auto-regressive text generation task to generate the relation between given objects. Therefore, in this work, we introduce image semantic relation generation (ISRG), a simple but effective image-to-text model, which achieved 31 points on the OpenPSG dataset and outperforms strong baselines respectively by 16 points (ResNet-50) and 5 points (CLIP).
\end{abstract}

\section{Introduction}
\label{sec:intro}
The PSG classification task aims to identify the three most salient relations in a given image~\cite{jingkang50}. Unlike generating a full scene graph~\cite{xu2020survey}, this task does not require the model to find the objects corresponding to the relations~\cite{jingkang50}. Therefore, if we can settle the scene graph generation task or identify the relationship between the given subject and object, we can solve the PSG classification task thoroughly.

In this work, we first employ a panoptic image segmentation algorithm~\cite{kirillov2019panoptic} from Detectron~2~\cite{facebookresearch} to map each pixel to its object, then pick the top-k objects with large area ratios, and finally provide these objects into a multi-model model to generate sequences, under the prefix tree constraint, until we get three different relation descriptions.

The experiment results show that the generative relation extraction model has obvious advantages compared with the multi-label classification model (ResNet, ViT, and CLIP). The combination of the visual encoder and the language decoder enables the model to learn more profound relational semantic concepts, which makes our model significantly outperform other baseline models.

In summary, we make the following explorations on the OpenPSG dataset:

\textbf{Multi-modal Image-to-text Model}
We introduce the vision encoder-decoder model~\cite{chen2020multimodal} in the PSG classification task. Leveraging the strong learning abilities of the pretrained vision encoder and language decoder, our model shows the superior semantic understanding capability spanning image and text.

\textbf{End-to-end Scene Graph Generation}
Compared with CLIP, which only exchanges multi-modal information between image features and texture features at the attention matrix, our method ISRG fuses image and text messages during each token generation process~\cite{radford2021learning}. This way, the text and the image can be fully integrated, encouraging the model to comprehend relation semantic concepts.

\textbf{Comparison of Model Performance}
In the experiment section, we compare several baseline models, including ResNet-50, ViT, CLIP, and our method ISRG. Our model outperforms other baseline models on the OpenPSG dataset by improving the lack of previous models.

\begin{figure*}
    \centering
    \includegraphics[width=\textwidth]{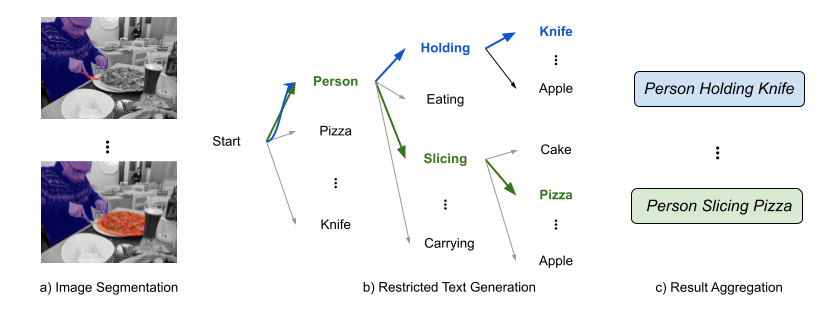}
    \caption{Example of a short caption, which should be centered.}
    \label{fig:short}
\end{figure*}

\section{Related works}
\label{sec:related}

\textbf{Scene Graph} is a graph structure \(G = (O, E)\) extracted from a given image, which consists of \(N\) object nodes \(O = (O_1, O_2, ... , O_n)\) and \(M\) relations between nodes \(E = (E_1, E_2, ... , E_m)\). Therefore, we can use triplets \(T = (O_{head}, E_{relation}, O_{tail})\) on this directed graph to describe relationships between two objects from a given image~\cite{xu2020survey}. For example, location relations like  ("zebra in front of elephant") or state relations like ("person holding baseball bat").

\textbf{Residual Neural Network (ResNet)} is an open-gated variant of HighwayNet~\cite{srivastava2015highway}. Due to the skip mechanism, ResNet is the first feed-forward neural network to reach hundreds of layers. The residual connection effectively simplifies the network while mitigating the degradation problem of deep networks. Since its introduction, the ResNet model family has always been used as a strong baseline to compare with new models~\cite{he2016deep}.

\textbf{Vision Transformer (ViT)} demonstrates that the dependence of computer vision on CNNs is dispensable~\cite{zhao2021battle}. Applying sequences of image patches directly to a pure transformer can also perform very well on computer vision tasks. Compared with state-of-the-art convolutional networks, ViT attains impressive results while requiring substantially fewer computational resources to train~\cite{dosovitskiy2020image}. In this paper, ViT refers to the ``vit-large-patch16-224" variant, unless otherwise stated.

\textbf{GPT-2} is a decoder-only transformer model pretrained on a tremendous amount of English corpora~\cite{radford2018language}. The model will auto-regressively generate subsequent tokens based on previous tokens. More precisely, as a sequence-to-sequence model, the model input is a sequence of text and the output is the same sequence but shifted one token to the right. 

\textbf{CLIP} is a multi-modal vision and language model which combine a ViT transformer and a causal language model transformer to obtain visual and text features separately. It aims to narrow the distance between positive pairs and keep antagonistic pairs as far away as possible by a contrastive learning approach of projecting textual and visual features into a latent space with the same dimension and calculating their dot product similarity scores~\cite{radford2021learning}. 

Regarding the PSG classification task, CLIP exhibits excellent zero-shot learning ability. With proper prompt guidance, frozen CLIP achieves performance comparable to the fine-tuned ViT model on the validation data split. However, since the CLIP training data includes the COCO dataset and 400 million extra data samples, such comparisons may be unfair~\cite{radford2021learning}.

\section{Problem and Dataset}
\label{sec:problem_dataset}
\textbf{PSG Generation} groups pixels into different nodes based on its pixel-granular image segmentation, and then predicts the relationship between these nodes~\cite{yang2022panoptic}. Different from the classic scene graph generation task, the panoptic scene graph generation task uses image segmentation instead of rigid bounding boxes~\cite{li2017scene, xu2020survey}. This requires the model need to understand the relative relationship between objects in more detail.

\textbf{PSG Classification} is a simplified version of the PSG Generation. In lieu of identifying the objects, the model needs to output the three most salient relations in the given images~\cite{jingkang50}. The relations are from a dedicated list, which includes 56 relation classes. Since the first 6 relations (over, in front of,  beside, on, in, attached to) are too ambiguous, they have been removed from this task. Therefore, the PSG classification is a 50-class multi-label classification task~\cite{jingkang50}.

\textbf{Datasets} for benchmarking, we are using the OpenPSG dataset~\cite{yang2022panoptic}, which contains 49k well-annotated images from COCO~\cite{lin2014microsoft} and Visual Genome~\cite{krishna2017visual}.

\section{Methodology}
\label{sec:method}
Although the original PSG classification task~\cite{jingkang50} does not require associating predicted relationships with their corresponding objects, we believe that it is more intuitive to predict relationships after presenting objects. Therefore, we decompose the original task into three parts: 1) Panoptic Segmentation and Subject Selection 2) Restricted Image-to-Text Generation 3) Relation Aggregation. In this section, we first introduce the model components separately, then show how these components work together, and finally describe model training optimization and regularization methods.

\subsection{Panoptic Segmentation and Subject Selection}
\textbf{Panoptic Segmentation} To extract object information from the COCO dataset, we adopt panoptic segmentation to all the images in the dataset. The panoptic segmentation task is a strict generalization format of semantic segmentation. Given a predetermined \(N\) set of semantic classes \(C := {C_0, . . . , C_{n-1}}\), the task requires mapping each pixel \(i\) from an image to a pair \((C_i, I_i) \in C \times N\), where \(C_i\) represents the semantic class, and \(I_i\) represents its instance id.

\begin{figure}[]
  \centering
  \includegraphics[width=\linewidth]{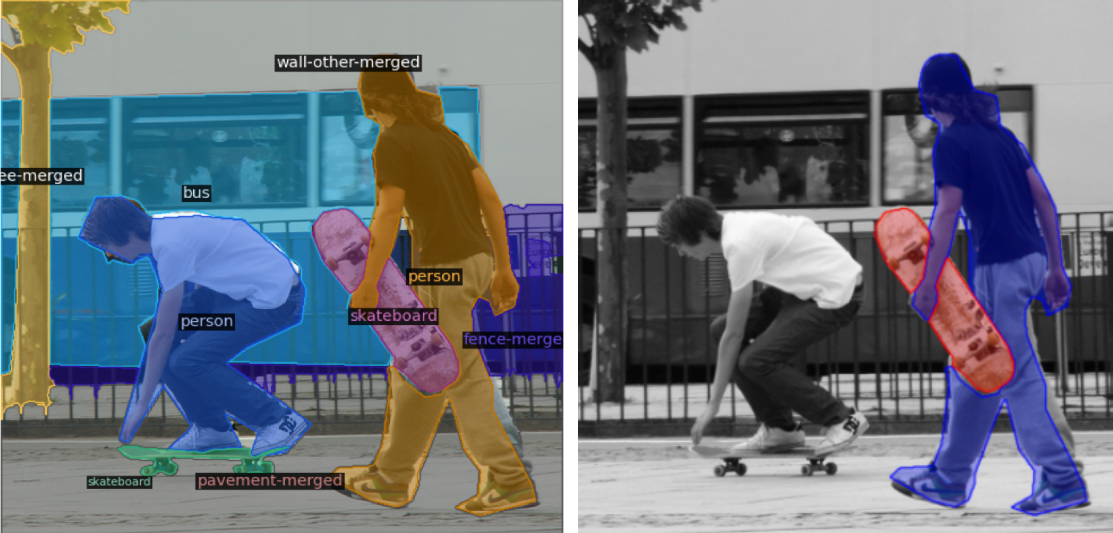}

   \caption{Examples of image segmentation.
   The left image is the conventional panoramic segmentation result, and the right image is the input image for predicting the relationship between the subject (red) and the object (blue).}
   \label{fig:segmentation}
\end{figure}

We employed a pre-trained model \footnote{COCO-PanopticSegmentation/panoptic\_fpn\_R\_101\_3x.yaml} in the Detectron~2~\cite{facebookresearch} platform to handle the image segmentation work. Figure~\ref{fig:segmentation} illustrates an example of the input and output of the image segmentation model.

\textbf{Subject Selection} During the training phase, we collect all relation triples in the PSG dataset to generate training sample pairs \mbox{\(<image, text>\)}, where \(image\) highlights the corresponding subject (red) and object (blue), and grayscales other areas, and \(text\) is composed of \mbox{\(<subject, relation, object>\)}. For example, the text corresponding to the right image of Figure~\ref{fig:segmentation} is ``person holding skateboard".

For the inference phase, traversing all relations between objects consumes excessive computational resources. Therefore, we only select the first five objects with a large area ratio for analysis.

\subsection{Restricted Image-to-Text Generation}
Inspired by the image captioning task~\cite{vinyals2016show}, we create a hybrid model jointing a ViT encoder with a GPT-2 decoder, which takes an image as input and generates a textual description containing relations. Unlike CLIP, which only mixes image and text embeddings in the matrix dot product stage~\cite{radford2021learning}, our model mingles image and text embeddings in each token generation to make the information exchange more adequate. 

However, the conventional autoregressive generation cannot guarantee that the generated results are entirely consistent with the relation candidates, so we import prefix tree to restrain the outcome~\cite{de2020autoregressive}. Assuming that the PSG training set contains all possible object-relation triples, we first employ the CLIP tokenizer to encode all object-relation triples into lexical ids sequences, and add them to the prefix tree one by one, then we can get a prefix tree containing all possible output sequences. Specifically, when calculating the probability \(P(T)\), supplemental restrictions on the candidate set \(S(T_n)\)need to be added according to the prefix tree, where:

\begin{equation}
P(T) = \prod_{n=1}^{L} P(T_{n}|T_{1}, ..., T_{n-1})
\end{equation}

\begin{equation}
S(T_{n}) = Prefix\_Tree(T_{1}, ..., T_{n-1})
\end{equation}

\begin{equation}
T_{n} \in S(T_{n})
\end{equation}

By setting a reasonable length of beam search, supplementing with the prefix-tree, restricted autoregressive generation mechanism can find the optimal object relation triples and avoid irrational consequences, such as "elephant on the back of person" or "train driving on water".

\subsection{Relation Aggregation}
Since the PSG classification task does not require predicting objects corresponding to relations, we correlate the probability of generated sequences \(P(T)\) as the probability of relations. In the inference stage, we sample as many sequences as possible for each image until we get three distinct relations.

\subsection{Optimisation}
\textbf{Optimiser} 
To properly introduce weight decay and speed up loss function convergence, the model uses AdamW~\cite{loshchilov2017decoupled} as the optimiser.

\textbf{Learning Rate}
We employ the Bayesian hyper-parameter search method~\cite{xia2017boosted} to determine the best learning rate in the range from 1e-8 to 1e-5, and finally adopt 1e-6 as the initial learning rate.

\textbf{Weight Decay}
Because AdamW introduces the concept of momentum, weight decay in AdamW is not equivalent to L2 regularisation~\cite{loshchilov2017decoupled}. We attempted a set of values [1-e2, 1-e3, 1-e4, 1-e5], and finally settled on 1-e3.

\textbf{Loss Function} Here we inherit the Cross Entropy Loss function from GPT-2 model with a mean weight reduction.

\textbf{Batch Size} is 32, which is manually selected from a list [16, 32, 48, 64].

\textbf{Training Epochs} We set a maximum of 48 training epochs, although models typically converge around 30 epochs.

\textbf{Machine Configuration} All experiments were run on 4~*~A100~SXM4.

\subsection{Regularisation}
\textbf{Dropout}
We keep all dropout terms in GPT-2 model and increase ``hidden\_dropout\_prob" and ``attention\_probs\_dropout\_prob" in ViT model from 0.1 to 0.2.

\textbf{Data Augmentation}
At the beginning of the project, we tried to use RandomHorizontalFlip and ColorJitter to enhance the training data, but later we realised that ColorJitter would cause the colour of the highlighted markers to change. So in the final version of the data loader, we only used RandomHorizontalFlip.

\section{Experiments}
\label{sec:experiments}

\subsection{Main Results}
Table~\ref{tab:main_result} lists model performance on the OpenPSG test dataset~\footnote{The matriculation number is G2204045F}, where mean recall is used as the evaluation metric. Except our model ISRG, all other models follow the classification manner.

\begin{table}[]
    \caption{Model Performance on the OpenPSG Test dataset}
    \centering
    \begin{tabular}{llc}
        \hline
        Model            & Type           & Mean Recall         \\ \hline
        ResNet-50        & Classification & 15.50               \\
        ViT              & Classification & 23.49               \\
        CLIP             & Classification & 26.60               \\
        ISRG (Our Model) & Generation     & \textbf{31.00}      \\ \hline
    \end{tabular}
    \label{tab:main_result}
\end{table}

\subsection{Ablation Study}
    \begin{table}[]
        \caption{List of Ablation Experiments}
        \label{tab:ablation}
        \centering
        \begin{tabular}{llc}
            \hline
            Type         & Name                          & Mean Recall (\%)     \\ \hline
            RTG          & Unrestricted Model            & 6.19                 \\
            RTG          & Restricted Model              & \textbf{31.00}    \\ \hline
            OH           & Without Processing            & 22.43                \\
            OH           & Grey Processing               & 16.57                \\
            OH           & Random Colour Highlight       & 21.24                \\
            OH           & Specific Colour Highlight     & \textbf{31.00}    \\ \hline
            OS           & Select Top 1 Subject          & 26.06                \\
            OS           & Select Top 3 Subjects         & 28.06                \\
            OS           & Select Top 5 Subjects         & \textbf{31.00}    \\
            OS           & Select Top 7 Subjects         & 30.09                \\
            OS           & Select All Subjects           & 29.89                \\ \hline
        \end{tabular}
    \end{table}

    \textbf{Restricted Text Generation (RTG)} 
    A prefix tree provides constraints during the textual autoregressive generation. By comparing the results of the restricted generative model and the unrestricted generative model on the OpenPSG test dataset, we observe that although the unrestricted generative model can also generate sequences that are semantically similar to ground truth labels after being fine-tuned on the OpenPSG training dataset, the generated sequence cannot exactly match the given relation set. From table~\ref{tab:ablation}, ablation tests show that the unrestricted generative model can hardly perform well on the OpenPSG task, while under the constraints of the prefix tree, the generative model can achieve considerable results.
    
    \textbf{Object Highlight (OH)} 
    We hypothesize that highlighting the subject and object on the input images can help the model perceive the given subject and object in generating text sequences. As shown in table~\ref{tab:ablation}, we tried not applying any processing on the input image, grey-processing all pixels except the subject and object, highlighting the subject and object with random colours, and annotating the subject and object with a specific colour~(red \& blue). Among these experiments, the scheme using specific colour annotations got the best result, proving that highlighting objects can improve the model's performance.
    
    \textbf{Object Selection (OS)}
    To improve the cost-effectiveness of computing resources, only objects with a larger area proportion are selected for subsequent text sequence generation. Hence, we design an series of ablation experiments to test the effectiveness of this subject selection method. In Table~\ref{tab:ablation}, these experiments demonstrate that the model performance has already reached the same level of traversing all objects when selecting more than three objects for the text generation.

\section{Conclusion}
We reformulated the PSG classification task pipeline by introducing an image-to-text transformer model and a restricted text generation scheme. Based on the new formulation, we propose a simple and effective relation extraction model ISRG, which illustrates considerable advantages to previous works and provides a novel manner to construct scene graphs, and serves as a strong baseline.

After designing and implementing the model pipeline, we realized that a filling generation, such as "Person \{relation\} Pizza" and "Train driving on \{object\}", explicitly provides subject and object information in text features, which might help the model better understand semantic relation concepts. We leave it for further study.

{\small
\bibliographystyle{ieee_fullname}
\bibliography{egbib}
}

\end{document}